\documentclass[authoryear,preprint,review,12pt]{elsarticle}

\usepackage{amssymb}
\usepackage{amsmath}

\usepackage{multirow}
\usepackage{booktabs}
\usepackage{pifont}
\usepackage{url}
\usepackage{tcolorbox}
\usepackage{subcaption}

\usepackage{algorithm}
\usepackage{algpseudocode}
\usepackage{setspace}

\newcommand{\cmark}{\ding{51}}
\newcommand{\xmark}{\ding{55}}

\begin{document}

\begin{frontmatter}



\title{From Entity-Centric to Goal-Oriented Graphs: Enhancing LLM Knowledge Retrieval in Minecraft}

\author[angel]{Jonathan Leung\corref{cor1}} 
\cortext[cor1]{Corresponding author.}
\ead{jonathan.leung@ntu.edu.sg}
\author[angel]{Yongjie Wang}
\author[ccds]{Zhiqi Shen}
\affiliation[angel]{organization={Alibaba-NTU e-Sustainability CorpLab (ANGEL), Nanyang Technological University},
            addressline={50 Nanyang Avenue}, 
            postcode={639798}, 
            country={Singapore}}

\affiliation[ccds]{organization={College of Computing and Data Science, Nanyang Technological University},
            addressline={50 Nanyang Avenue}, 
            postcode={639798}, 
            country={Singapore}}


\begin{abstract} 
Large Language Models (LLMs) demonstrate impressive general capabilities but often struggle with step-by-step procedural reasoning, a critical challenge in complex interactive environments. While retrieval-augmented methods like GraphRAG attempt to bridge this gap, their fragmented entity-relation graphs hinder the construction of coherent, multi-step plans. In this paper, we propose a novel framework based on Goal-Oriented Graphs (GoGs), where each node represents a goal and edges encode logical dependencies between them. This structure enables the explicit retrieval of causal reasoning paths by identifying a high-level goal and recursively retrieving its prerequisites, forming a coherent chain to guide the LLM. Through extensive experiments on the Minecraft testbed, a domain that demands robust multi-step planning and provides rich procedural knowledge, we demonstrate that GoG substantially improves procedural reasoning and significantly outperforms GraphRAG and other state-of-the-art baselines.
\end{abstract}






\end{frontmatter}



\section{Introduction}

Large Language Models (LLMs) have recently been applied as reasoning and planning components in interactive environments, where they enable dynamic decision-making for agents such as non-player characters (NPCs) and virtual assistants. Games, in particular, have become valuable testbeds for studying the reasoning capabilities of LLMs because they combine structured rules with open-ended objectives~\citep{gallotta2024large}. While early research has explored strategic domains like chess~\citep{NEURIPS2023_16b14e3f}, the frontier has moved toward open-world settings that demand long-horizon, hierarchical goal decomposition. Environments like Minecraft~\citep{wang2024voyager,zhu2023ghostminecraftgenerallycapable} have emerged as critical benchmarks for this challenge due to their combinatorial action space and the need for multi-step procedural reasoning.

To ground LLMs in such complex domains, Retrieval-Augmented Generation (RAG) has become a standard approach~\citep{gao2023retrieval}. State-of-the-art methods like GraphRAG~\citep{edge2025localglobalgraphrag} structure external knowledge into entity-relation graphs to facilitate retrieval. However, this entity-centric paradigm is fundamentally ill-suited for procedural tasks. It fragments causal knowledge into an excessive number of low-granularity triples, making it difficult to reconstruct a coherent, step-by-step plan. This is not just a theoretical issue; in our experiments, this fragmentation introduces significant noise that hinders the agent's performance.

\begin{figure}
    \centering
    \includegraphics[width=0.8\linewidth]{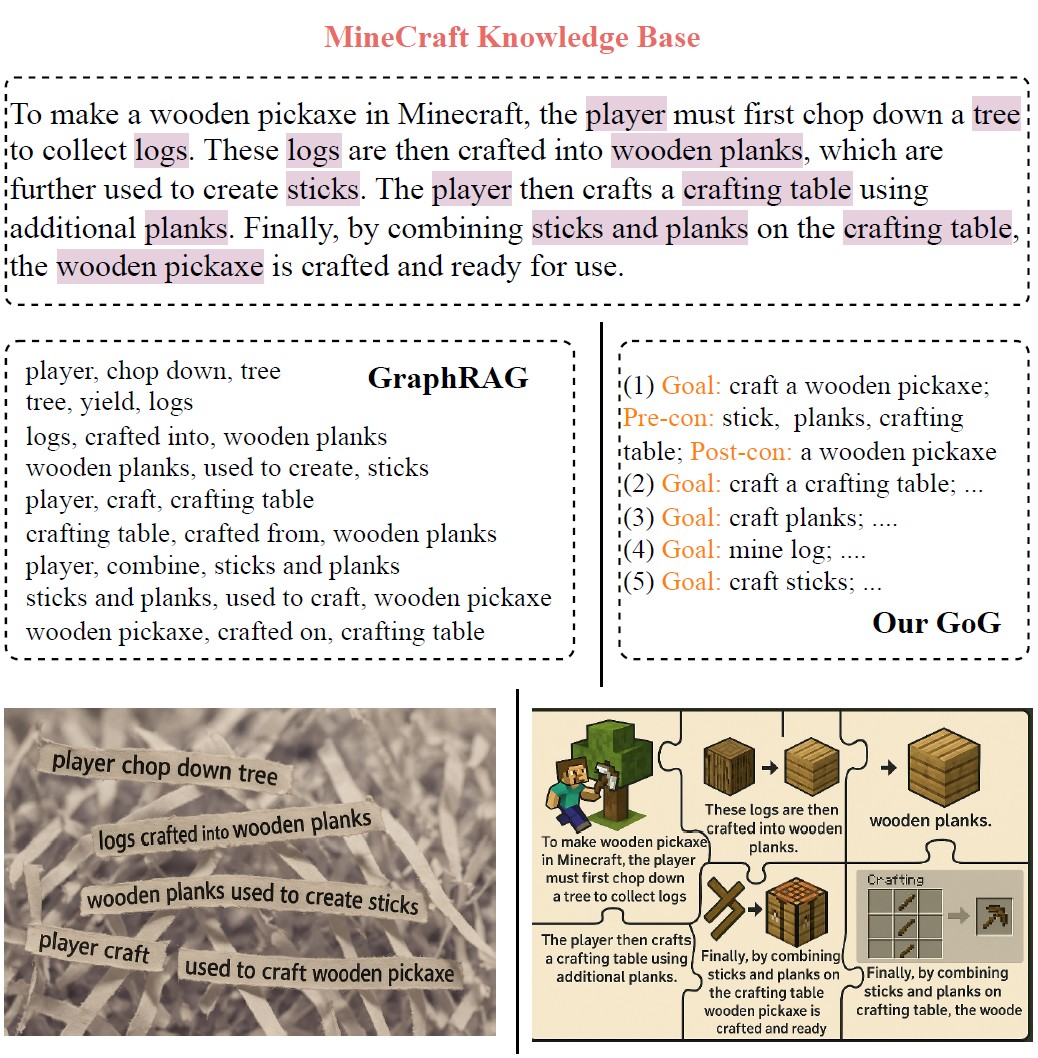}
    \caption{GraphRAG extracts an excessive number of low-granularity entity–relation triples, which hinders effective reasoning over fragmented information. In contrast, our GoG captures procedural knowledge through goal hierarchies, thereby supporting coherent and structured reasoning.}
    \label{fig:motivation}
\end{figure}

Reconstructing a coherent plan from this fragmented knowledge is akin to the saying, ``tearing paper is easy, putting it back together is hard'', illustrated in Figure~\ref{fig:motivation}. This motivates our work: to design a knowledge framework that explicitly captures procedural dependencies. We introduce the Goal-Oriented Graph (GoG), a novel structure where nodes represent goals and directed edges encode the prerequisite relationships between them. Using this graph, our retrieval process identifies a high-level goal and recursively retrieves its subgoals, forming a coherent reasoning chain to guide the LLM. 

We focus on Minecraft as a deep and open-ended benchmark for multi-step reasoning rather than as a generic game environment. While the GoG framework is conceptually general, in this work we explicitly target domains where procedural knowledge can be externalized into goal–precondition structures, and we do not claim applicability to arbitrary game environments. This focused scope allows us to rigorously evaluate how goal-oriented knowledge retrieval improves LLM reasoning in a well-studied, reproducible setting.

Our contributions are summarized as follows:
\begin{itemize}
\item We introduce Goal-Oriented Graphs (GoGs), a novel framework designed to enhance the procedural, multi-step reasoning of LLMs. GoG is designed to leverage external procedural descriptions that can be organized into goals, preconditions, and postconditions, a common characteristic of instructional and technical domains. GoGs model how complex tasks decompose into actionable subgoals, shifting the paradigm from entity-centric relations to logical goal dependencies.
\item We propose a goal-driven retrieval algorithm that traverses the GoG to construct coherent and explicit reasoning chains, overcoming the fragmented retrieval of traditional graph-based methods for procedural tasks.
\item We validate our framework through extensive experiments in Minecraft, a demanding benchmark for long-horizon procedural planning, demonstrating its effectiveness in Minecraft and, more broadly, in domains characterized by structured, goal-oriented procedural knowledge.
\end{itemize}

\section{Related Work}
\textbf{LLM-Based Agents} have been proposed for Minecraft. For example, Voyager is an agent that learns skills via lifelong learning~\citep{wang2024voyager}. MP5~\citep{qin2024mp5} focuses on leveraging multimodal LLMs to perform planning based on what the agent sees. Unlike these methods, our method aims to improve an agent's performance by using goal-oriented knowledge extracted from text sources. Two methods that are more closely related to ours are GITM~\citep{zhu2023ghostminecraftgenerallycapable} and Optimus-1~\citep{li2024optimus}. GITM uses text knowledge from the Minecraft Wiki and in-game recipes, as does our method. However, we focus on the construction of Goal-Oriented Graphs from text sources and the retrieval process for planning. On the other hand, GITM is presented as a unified agent system.
Optimus-1 is a Minecraft agent that contains a hierarchical knowledge graph that is used to decompose goals into subgoals. However, the process by which their graph is created is different from ours as it only uses recipes from the game files, and their agent is focused on the utilization of multimodal memory.

\textbf{Reasoning using LLMs} is a popular research area that aims to enable LLMs to handle more complex tasks. There are a wide variety of methods, including those that alter the prompt to encourage the LLM to output intermediate steps~\citep{wei2022chain}, provide feedback from the environment to the LLM so that it can adjust its behaviour accordingly~\citep{yao2023react,shinn2023reflexion}, maintain and expand multiple generated reasoning sequences~\citep{yao2023tree}, and use majority choice voting~\citep{wang2023selfconsistency}. 

\textbf{RAG-Based Systems} aim to address LLMs' knowledge gaps. RAG maintains a vector database of source documents that can be retrieved and given to the LLM as context information according to the embedding similarity of the source documents to a given query~\citep{NEURIPS2020_6b493230}. However, if the answer to a given query spans across several documents, then the retrieval process may fail to contain the answer. Therefore, GraphRAG and its variants were proposed to address this issue~\citep{edge2025localglobalgraphrag,wu2024medical,guo2024lightrag}.

\textbf{Hierarchical Decomposition} is a traditional method in AI to break complex problems into smaller sub-problems that has been used in areas including planning~\citep{ghallab2004automated,erol1994htn} and reinforcement learning~\citep{dietterich1998maxq,sutton1999between}. These methods typically rely on domain knowledge or hand-crafted rules to establish the hierarchy of tasks. In this work, we aim to construct a hierarchy of goals in the form of a graph by constructing it from text data sources.

\section{Methodology}

\begin{figure}
    \centering
    \includegraphics[width=1.0\linewidth]{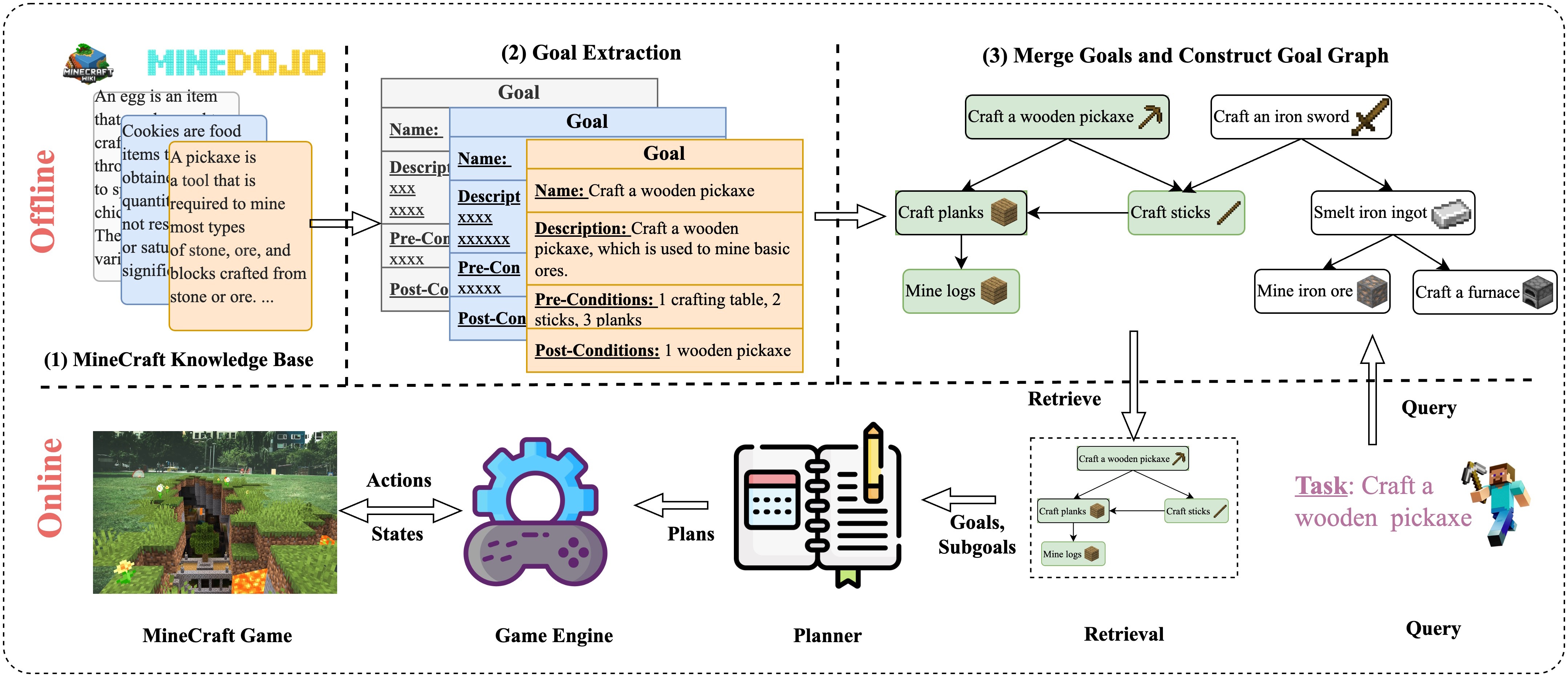}
    \caption{An overview of our proposed method GoG. First, we construct a knowledge base of goals from source text documents. Then, given a task instruction, we retrieve goal-oriented knowledge from the knowledge base to use for plan generation.}
    \label{fig:method_overview}
\end{figure}

Our method, illustrated in Figure~\ref{fig:method_overview}, consists of two main phases: (1) goal-oriented knowledge base construction and (2) reasoning-aware inference. In the first phase, we construct a directed graph whose nodes represent goals such as ``craft a wooden pickaxe'' and edges represent subgoal relationships between goals. To do this, we use an LLM to extract structured goal information from textual sources such as Minecraft Wiki pages and recipe descriptions.
During inference, we use the constructed knowledge base to recursively search for step-by-step reasoning paths, which are followed to guide LLM prompting for the given task.

A key design decision in our framework is the integration of classical goal-structured reasoning with LLM-based semantic grounding. Although the subgoals extracted from the GoG form a structured, planner-like hierarchy, a purely symbolic planner would still require a domain-specific, hand-engineered model to translate abstract goals into executable actions. In contrast, we employ the LLM both during graph construction, to extract and represent knowledge from text, and during inference, to ground structured goals into natural language plans executable by a downstream controller, which in this case is STEVE-1~\citep{lifshitz2023steve}.

This design reflects a synergistic hybrid architecture:
\begin{itemize}
    \item GoG provides the formal reasoning structure: a high-level causal plan ensuring logical correctness between goals and subgoals.
    \item The LLM serves as both a knowledge extractor and a semantic translator, converting unstructured text into structured goal graphs and grounding those abstract dependencies into executable action descriptions.
\end{itemize}

This hybrid design aligns with the emerging ``language-as-action'' paradigm in modern agent systems, such as SayCan~\citep{ahn2022can} and STEVE-1, where language serves as the universal interface between reasoning and control.

In the following subsections, we describe the two phases of our proposed method in detail, using Minecraft as the testbed for illustration. Details on the computational cost of both phases are provided in \ref{appendix:construction}.

\subsection{Goal Knowledge Base Construction}
In the first phase of our method, we construct a directed graph $G = (V,E)$, where $V$ is a set of nodes and $E \subseteq V \times V$ is a set of edges from source to target nodes. The algorithm is shown in Algorithm~\ref{alg:gog_construction}, and consists of three main steps: goal extraction, goal merging, and subgoal derivation.

\begin{algorithm}[H]
\begin{spacing}{0.9}
\small
\caption{Goal-Oriented Graph (GoG) Construction}
\label{alg:gog_construction}
\begin{algorithmic}[1]
\Require 
    \textit{Corpus}, \textit{LLM}, Embedding function $f$, Threshold $\theta$
\Ensure 
    $G = (V, E)$: A Goal-Oriented Graph
\State $G \gets (V \gets \emptyset, E \gets \emptyset)$
\For{each chunk $c_i \in \text{SplitCorpusIntoChunks}(\textit{Corpus})$}
    \State $G_{ext} \gets \text{LLM.ExtractGoals}(c_i)$
    \For{each new\_g $\in G_{ext}$.Nodes}
        \State existing\_g $\gets \underset{g \in V}{\text{argmax}} \text{ sim}(f(\text{new\_g.name}), f(g.\text{name}))$
        \State name\_sim $\gets \text{sim}(f(\text{new\_g.name}), f(\text{existing\_g.name}))$
        \State cond\_ok $\gets \text{CheckCondSim}(\text{new\_g, existing\_g}, f, \theta)$
        
        \If{cond\_ok and name\_sim $\geq \theta$}
            \Statex \Comment{\parbox[t]{\dimexpr\linewidth-2\algorithmicindent}{Equivalent goal based on high name and condition similarity. No action needed.}}
            \State continue
        \ElsIf{cond\_ok and name\_sim $< \theta$} \Comment{Alias goal.}
            \State Add new\_g.name to existing\_g.aliases
        \Else \Comment{New distinct goal.}
            \State $V \gets V \cup \{\text{new\_g}\}$
            \State DeriveSubgoalEdges($\text{new\_g}, G, f, \theta$)
        \EndIf
    \EndFor
\EndFor
\State \Return $G$
\end{algorithmic}
\end{spacing}
\end{algorithm}

\subsubsection{Goal Structure and Extraction}
Each node in the graph represents a goal, and each one is associated with a set of attributes consisting of its name, aliases, description, preconditions, and postconditions, which are defined in Table~\ref{tab:goal_definition}. The names of the nodes in our graph are goal-oriented phrases that succinctly describe the action involved and the expected outcome of the goal, which is detailed further in the node's description. The preconditions are a list of prerequisites that are needed before the goal can be pursued, and postconditions explicitly describe the expected outcome of achieving the goal. Edges represent subgoal relationships between goals, and each edge is associated with a description to briefly explain how two goals are related. 

\begin{table}[]
    \centering
    \begin{tabular}{l p{9cm}}
        \toprule
        Attribute & Definition \\
        \midrule
        Name & Short, specific labels of the form ``$<$action$>$ $<$minecraft\_item$>$'' (e.g., ``craft planks'', ``mine cobblestone'').\\
        Description & A concise natural language summary of the goal. \\
        Required Tools & Tools needed to accomplish the goal.\\
        Required Materials & Materials required for the goal.\\
        Postconditions & The outcome of completing the goal (e.g., newly crafted item). \\
        \bottomrule
    \end{tabular}
    \caption{Definitions of attributes of goals in GoG.}
    \label{tab:goal_definition}
\end{table}

In the case of Minecraft, to craft a wooden pickaxe, our constructed graph includes a set of goals such as “craft a wooden pickaxe” and “craft planks.” Since both planks and sticks are required, the graph contains directed edges from the “craft a wooden pickaxe” node to the “craft sticks” and “craft planks” nodes, as shown on the right side of Figure~\ref{fig:method_overview}. Furthermore, the goal “craft planks” depends on acquiring logs, introducing an additional subgoal relationship. In this manner, high-level goals are recursively decomposed into subgoals until they reach atomic operations. Preconditions, such as required tools and materials, are also encoded in the graph to represent dependencies for actions like crafting tools or mining ores.

To extract the goal-related information from large text-based data sources, we utilize LLMs. However, LLMs have a limit to the number of tokens that they can process in one query, and therefore the source text needs to be split into smaller chunks that can fit an LLM's context size. For each chunk, we use an LLM to extract goals, their attributes, and subgoal relationships between goals. The prompt used for goal extraction is provided in Figure~\ref{fig:goal_extraction_prompt}. 

\subsubsection{Goal Merge and Subgoal Derivation}

After extracting goals and subgoals from all text chunks, two critical challenges must be addressed: (1) duplicate goals may be extracted from different chunks, how can they be effectively merged? and (2) a goal identified in one chunk may have a subgoal relationship with a goal from another chunk, how can the complete subgoal relations be derived? 

The equivalence between new and existing goals is determined by the similarity between their name embeddings, and further verified using the preconditions and postconditions of each goal pair. To perform the aggregation process, we use a text embedding function $f: \mathcal{T} \rightarrow \mathbb{R}^d$ to map a given text $t \in \mathcal{T}$ to a $d$-dimensional vector. In particular, for each newly extracted goal, we retrieve the most similar goals from the existing goal base by comparing the cosine similarity between their name embeddings. Then, we compare the preconditions and postconditions of the pairs of new goals and their most similar existing goals. For a given new goal and its most similar goal, we sort each of their lists of preconditions and postconditions in alphabetical order, and then we calculate the cosine similarity between the embeddings of the pair after sorting.  We define a threshold to binarize the similarity of text embedding. 

If all pairs of corresponding conditions (preconditions and postconditions) have similarity scores above a threshold $\theta$, the new goal is considered equivalent to an existing one. Otherwise, the new goal is added to the goal base as a distinct entry. In cases where the condition similarity is high but the name embedding similarity is lower than $\theta$, we treat the new goal as an alias and append its name to the existing goal's alias list. If both condition and name similarities are high, the goals are treated as equivalent.

Once we have added a newly extracted goal to the knowledge base, the next step is to determine whether it has any subgoal relationships with any existing goals in our knowledge base. To do this, we use a process similar to the goal equivalence comparison described previously, but instead we will match preconditions of the goal to the postconditions of existing goals, and vice versa. If there is a match, we add a new edge to the knowledge base. After removing duplicate goals and completing subgoal relations across all chunks, we obtain a directed goal-oriented graph from the knowledge source, as shown in Figure~\ref{fig:method_overview}. 

\begin{figure}
    \centering
    \includegraphics[width=1.0\linewidth]{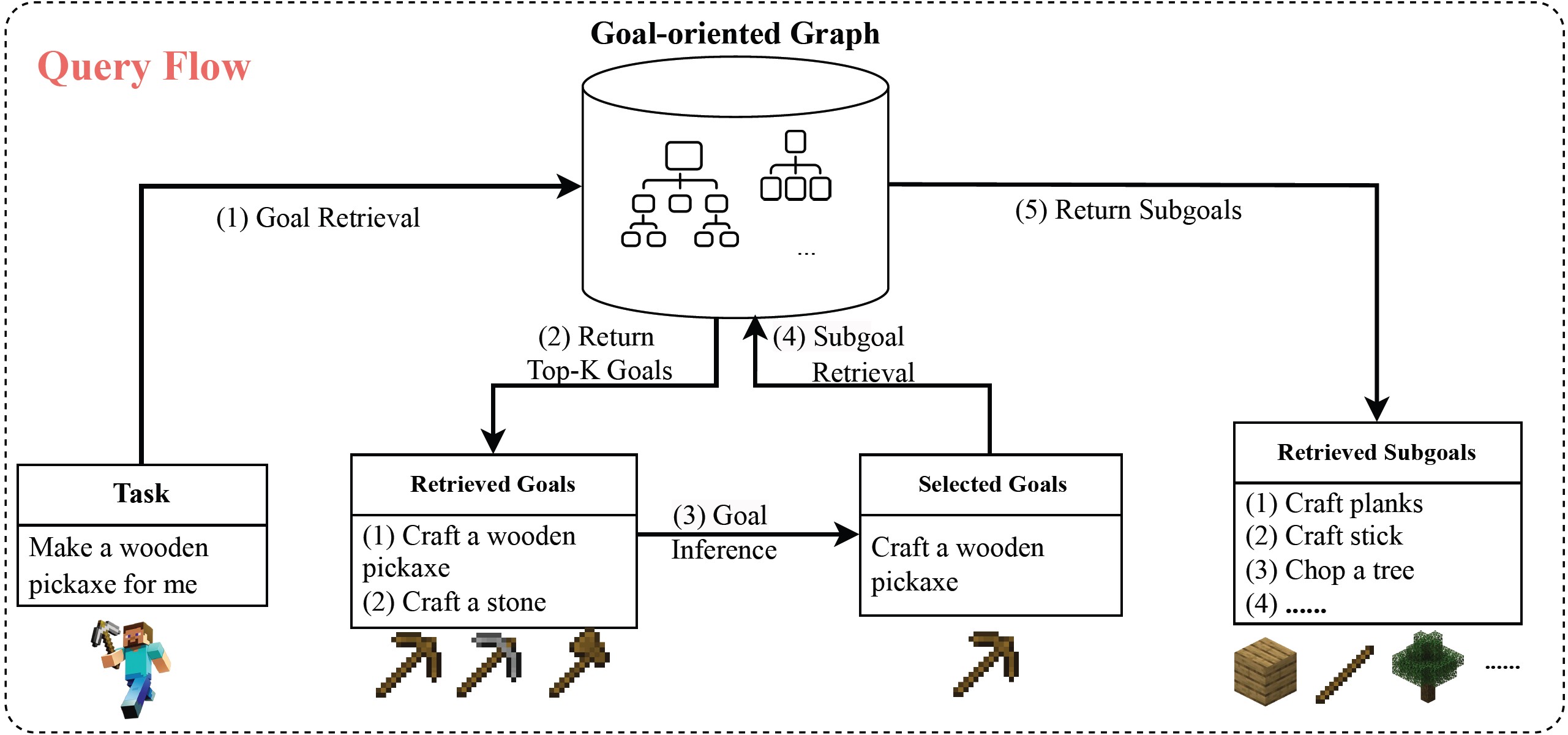}
    \caption{The query pipeline. For a given task, the top-$k$ goals based on embedding similarity between the query and the goals' names are retrieved.}
    \label{fig:query_pipeline}
\end{figure}

\subsection{Goal Selection and Planning}
\label{sec:goal_selection}

After constructing the knowledge base, we leverage it to extract goal-oriented knowledge for downstream tasks such as planning. The overall process is illustrated in Figure~\ref{fig:query_pipeline}. Given a task query, such as ``craft a wooden sword'', we retrieve the top-$k$ goals from the knowledge base according to the cosine similarity between the query and the goals' names in the knowledge base. If $k > 1$, we need to determine which goal is the best match for the query. To do this, we utilize LLMs to select the goal from the top-$k$ retrieved options by giving it the goal names, descriptions, and postconditions. The prompt given to an LLM to select the best matching goal from the $k$ candidates is provided in Table~\ref{fig:goal_inference_prompt}.

After selecting the goal that matches the query, we retrieve all subgoals using depth-first search (DFS) starting from the selected goal. We avoid infinite loops caused by cyclical subgoal relations by not going deeper when a node that has already been visited is seen again. After retrieving subgoals, it may be the case that there are multiple ways to achieve the overall goal. In this case, a procedure to select the best set of subgoals is needed, which may depend on the use case scenario. 

With the set of goals, their attributes, and edges indicating subgoal relationships, we end up with a goal tree. We traverse the tree to obtain a complete list of preconditions which, in Minecraft, is a list of materials and tools needed to achieve the overall goal. Finally, we perform planning by providing this information to an LLM and prompting it to generate a plan. The planning prompt we use is provided in Table~\ref{fig:planning_prompt}. 

\subsection{GraphRAG vs GoG}

In this section, we summarize the major differences between GraphRAG and our proposed GoG. Although GoG and GraphRAG both construct knowledge graphs to aid in downstream tasks, there are key differences between the graphs that are listed in Table~\ref{tab:gog_vs_graphrag}. GraphRAG organizes knowledge into fragmented, low-granularity entity–relation triples, which limits its ability to support coherent reasoning, much like trying to reconstruct evidence from shredded pieces of paper. In contrast, we explicitly model abstract goals and subgoals as graph nodes, preserving the task hierarchy and enabling multi-step reasoning. For example, our GoG consists of 703 nodes and 1,653 edges, whereas the knowledge graph constructed using GraphRAG contains 12,388 nodes and 18,347 edges from the same source. This more compact and structured goal graph is better suited for goal achievement and planning, as evidenced by the following results on the Minecraft testbed.

\begin{table}[]
    \centering
    \begin{tabular}{l p{4.5cm} p{4.5cm}}
        \toprule
        Aspect & GoG & GraphRAG \\
        \midrule
        Graph Nodes & Structured task goals (e.g., ``craft a wooden pickaxe''). & Informational entities (e.g., people, places, events) extracted from documents. \\
        \midrule
        Graph Edges & Explicit goal–subgoal dependencies based on prerequisites. & Semantic or contextual relationships between entities. \\
        \midrule
        Node Semantics & Represent verb-centric goals with associated preconditions and postconditions. & Represent noun-centric entities or named entities.\\
        \midrule
        Purpose of Graph & Guides procedural planning and multi-step reasoning. & Provides rich, factual context for question answering or generation. \\
        \bottomrule
    \end{tabular}
    \caption{A comparison between the knowledge graphs constructed and used by GoG and GraphRAG.}
    \label{tab:gog_vs_graphrag}
\end{table}

\section{Experiments}

\subsection{Experimental Setup}
\label{sec:exp_setup}

In this work, we use the Minecraft environment and tasks provided by \citet{li2024optimus} to develop and evaluate our method. There are 66 tasks that are categorized into 7 groups: wood, stone, iron, gold, diamond, redstone, and armor. The task groups, ordered from easiest to hardest, are wood, stone, iron, gold, and diamond. The redstone and armor groups contain tasks of mixed difficulty ranging from iron to diamond. More difficult tasks require longer plans to be generated and require the agent to find rarer materials. The agent has a limited number of in-game steps, defined based on the task's group, to complete the task. The complete list of tasks and details can be found in Table~\ref{tab:exp_tasks}. Additionally, our code is publicly available\footnote{\url{https://github.com/jleung1/gog_minecraft}}.

At the beginning of each experimental run, the agent starts with an empty inventory and is given an instruction as a string, such as ``craft a wooden sword''. Then, as described in the previous section, we retrieve the top-$k$ goals from our knowledge base and use an LLM to perform goal inference by selecting a candidate from the top-$k$. Based on the selected goal, we retrieve all subgoals for the goal. Many items in Minecraft admit multiple crafting routes, often involving reversible transformations (e.g., combining iron ingots into a block of iron and decomposing the block back into ingots). To avoid trivial loops and ensure a well-defined reasoning chain, we explicitly exclude cyclical paths during depth-first search. Because the agent starts with an empty inventory and no prior resources, such cyclic expansions cannot contribute to a feasible plan and are therefore pruned. As a result, the retrieval process produces a single acyclic goal chain for each task. In this work, we focus on evaluating the feasibility, correctness, and material sufficiency of this retrieved chain, rather than enumerating or optimizing over multiple alternative decompositions.

After retrieving the subgoals, we parse the resulting goal tree, which consists of the goal, its subgoals, and all their attributes, to produce a list of items and materials needed to accomplish the task. Then, we provide the goal tree and list of items to an LLM to produce a plan to accomplish the task. The prompt used for planning can be found in Figure~\ref{fig:planning_prompt}. The plan consists of a sequence of subtasks and expected items to obtain from each subtask. This plan is given to the agent, which then attempts to execute the plan in Minecraft. 

During execution of the plan, if a step fails due to missing tool or materials, the Minecraft environment provides feedback to the agent about the missing item and quantity. This triggers the agent to replan, which follows a similar procedure as the original planning step at the beginning of the trajectory. For GoG, the list of materials and tools for the missing item is calculated from the goal tree, which is then given to the LLM to produce a plan to obtain missing materials. All baseline methods are able to perform replanning, with HKG also generating a list of materials using its knowledge graph. The generated replanning steps are then inserted into the original plan.

\subsubsection{Baselines}
We compare our proposed method \textbf{GoG} to three baselines:
\begin{itemize}
    \item \textit{Vanilla}. This method uses few-shot examples of tasks and directly generates plans for given tasks from a given task instruction, without any further context information. 

    \item \textit{GraphRAG}. We construct a knowledge graph from the Minecraft Wiki pages provided by MineDojo~\citep{fan2022minedojo} and in-game recipes using the method proposed by~\citet{edge2025localglobalgraphrag} using the same LLM (GPT-4o-mini) as GoG to ensure parity and isolate the impact of graph structure rather than model capability. During goal inference and planning, context information is retrieved using local search and inserted into the prompt given to the LLM. This method uses the same source as our GoG. The prompt used to extract entities and relationships is provided in Figure~\ref{fig:graphrag_prompt}, and other GraphRAG prompts were kept as the default.
    
    \item \textit{Hierarchical Knowledge Graph (HKG)}. This baseline represents a goal-oriented agent inspired by the Optimus-1 framework~\citep{li2024optimus}. It constructs a hierarchical graph from in-game recipe files, where each node corresponds to an item and edges capture prerequisite relations between items. For a given task, the agent identifies the corresponding node, extracts the list of required materials and tools, and uses an LLM to produce a plan. To ensure a controlled comparison focused solely on knowledge representation, we exclude Optimus-1’s additional multimodal components (memory and vision modules) and retain only the hierarchical knowledge reasoning mechanism.
\end{itemize}

Crafting recipes in the Minecraft Wiki are displayed as images. Therefore, we supplement the Wiki pages with recipes from the Minecraft game files to provide a text-based representation of the recipes. We remove pages that are unrelated to our experimental setting, such as those about real people related to the game, or game patch notes. Additionally, we filter pages by including those that have titles including the names of items contained in the in-game recipe files, or if the title is included in the name of the item. We only include text from the Wiki pages, leaving out images associated with the pages. The final set of documents used to construct our knowledge base consists of 514 Minecraft Wiki pages and 859 recipe files. Table~\ref{tab:exp_corpus} summarizes the corpora used by each method in our experiments.

It is worth noting that Minecraft recipes are inherently semi-structured and closely resemble goal–subgoal representations. This naturally benefits HKG, which directly relies on recipe dependencies. However, GoG extends beyond these predefined structures by integrating additional logical and procedural relations extracted from unstructured text sources such as Wiki pages, such as the type of tool needed to obtain a certain type of material, which are not captured in recipe trees.

To ensure a fair evaluation, both GoG and GraphRAG were constructed using the same knowledge sources (Minecraft Wiki and recipe files) and the same underlying model (GPT-4o-mini). This consistency isolates the contribution of graph representation and retrieval structure, rather than any differences in model or data quality.

\subsubsection{Models}
We utilize three LLMs in our experiments to balance statistical rigor with computational feasibility: Llama 3.2-Vision 90b, Gemma 3 27b, and Qwen 2.5-VL 32b. For our main comparative analysis, we employ Llama 3.2-Vision and Gemma 3. These experiments are computationally intensive, involving 30 independent runs for each task to ensure robust statistical comparisons between GoG and the baselines. By focusing this rigorous evaluation on two powerful and widely-used models, we can draw reliable conclusions about the performance of each knowledge framework.

To demonstrate the broader applicability of our approach, we include Qwen 2.5-VL, a model with a distinct architecture, in our ablation studies. This allows us to test the versatility of the GoG framework in a more targeted manner without incurring the prohibitive cost of running all models across all 30-run evaluations. We use a temperature of 0, context length of 32K, $\theta = 0.92$ with nomic-text-embed-v1.5~\citep{nussbaum2024nomic} as the embedding model, and use $k=3$ for both GraphRAG and our method. We provide game frames to the LLMs during goal inference and planning.

\begin{table}[]
    \centering
    \begin{tabular}{l|l}
         \hline
         Method & Corpus \\
         \hline
         Vanilla & None \\
         HKG & Recipe files \\
         GoG and GraphRAG & Minecraft Wiki and recipe files \\
         \hline
    \end{tabular}
    \caption{A summary of the corpuses used to construct the graphs in each method in our experiments.}
    \label{tab:exp_corpus}
\end{table}

\subsubsection{Evaluation metrics}
For our main experiments, we report the average \textit{success rates} of completing tasks and \textit{the number of in-game steps} used by the agent over 30 runs of each task for each combination of baseline method and LLM. Higher success rates and lower required steps indicate better performance, and the average steps include those from failed runs. This is so that a given model is not rewarded for failing on all the difficult, long-horizon tasks while only succeeding on the simple, short-horizon ones. This would lower the model's average steps if the metric were conditional on success, which may overstate the model's performance.

We conduct an ablation study in Section~\ref{sec:plan_ablation} and use metrics based on classical planning literature to assess the quality of the plans generated~\citep{ghallab2004automated}. We use goal satisfaction, soundness, completeness, and efficiency, which are defined in Section~\ref{sec:plan_ablation}, and are calculated using information from the retrieved goal tree for a given task. Using these metrics instead of success rate and in-game steps allows us to focus on the differences in generated plans without needing to consider other factors that may cause the agent to fail, such as the stochasticity of the environment. For all our experimental results, we report confidence intervals and identify the best performing models using pairwise t-tests using an alpha of 0.05.

After we generate a plan using the baselines and our method, we adopt STEVE-1 to convert text instructions into keyboard and mouse controls~\citep{lifshitz2023steve}. However, STEVE-1 is incapable of directly performing complex tasks that require multiple steps, hence the need for a planner that can decompose complex tasks into simpler ones.

\subsection{Main Results} 

The main results in Table~\ref{tab:main_results} show that GoG performs much better than the baseline models in more complex task groups. For simpler task groups (e.g., wood and stone), all methods achieve comparable performance in terms of success rate and the average number of steps required to complete the tasks. However, for more challenging task groups, such as iron, gold, and armor, our GoG demonstrates a significant advantage over the baselines. For example, the success rate on the gold task is three times higher than that of HKG, and on the armor task, GoG outperforms HKG by approximately 58\% with Llama 3.2-Vision. For the most difficult tasks, such as gold and diamond, all baseline methods consistently fail to achieve the goal within the maximum allowed in-game steps. In contrast, our method generally maintains a success rate above 50\%, demonstrating its robustness in long-horizon planning scenarios. These successful results empirically demonstrate that our GoG enhances the LLM's reasoning capabilities by providing goal structures and material lists that fill knowledge gaps and support accurate planning.

A key observation from our experiments is a notable pitfall of GraphRAG: it sometimes performs even worse than the Vanilla baseline, despite having access to additional retrieved content. The retrieval process of GraphRAG returns many irrelevant but connected nodes from the 1-hop neighbourhood. For example, a general entity such as stone can be used to craft many different tools. When tasked with making a stone axe, the LLM often lacks the reasoning to select the appropriate objects from its neighbourhood and instead returns all connected entities. This results in highly noisy context, which can significantly degrade downstream task execution. In this example of crafting a stone axe, the community reports of GraphRAG fail to translate into practical benefits, even with additional computational costs for clustering and summarization. This outcome echoes the old saying: reconstructing evidence from shredded pieces remains inherently difficult. 

\begin{table*}[!t]
    \centering
    \caption{The results of our main experiments, collected over 30 runs of each task for each combination of baseline method and LLM. Confidence intervals are calculated using an alpha of 0.05. Bolded numbers are the best result(s) of each task group that are significantly better than other methods as determined by pairwise t-tests.}
    \label{tab:main_results}
    \scalebox{0.5}{
    \begin{tabular}{l|c| cccc | cccc } 
    \toprule
    \multirow{2}{*}{Group} & \multirow{2}{*}{Metric}  & \multicolumn{4}{c|}{Llama 3.2 Vision} & \multicolumn{4}{|c}{Gemma 3} \\ \cmidrule{3-10}
      &  &  Vanilla & GraphRAG & HKG & GoG &  Vanilla & GraphRAG & HKG & GoG \\ \midrule 
     \multirow{2}{*}{Wood} & SR$\uparrow$ & 83.7 $\pm$ 4.2  & 80.0 $\pm$ 4.6 & \textbf{93.7 $\pm$ 2.8} & \textbf{95.7 $\pm$ 2.3} & 88.7 $\pm$ 3.6 & 91.0 $\pm$ 3.3 & \textbf{95.7 $\pm$ 2.3} & \textbf{93.3 $\pm$ 2.8} \\
     & AS$\downarrow$ & 1290 $\pm$ 70 & 1500 $\pm$ 70 & \textbf{1000 $\pm$ 60} & \textbf{1030 $\pm$ 50} & 1270 $\pm$ 70 & 1080 $\pm$ 70 & \textbf{990 $\pm$ 50} & \textbf{1050 $\pm$ 60} \\ \midrule 
     \multirow{2}{*}{Stone} & SR$\uparrow$ & 57.4 $\pm$ 5.9 & 37.8 $\pm$ 5.8 & 46.7 $\pm$ 6.0 & \textbf{80.0 $\pm$ 4.8} & 47.4 $\pm$ 6.0 & 47.8 $\pm$ 6.0 & \textbf{71.1 $\pm$ 5.4} & \textbf{69.6 $\pm$ 5.5}\\
     & AS$\downarrow$& 3900 $\pm$ 230 & 4610 $\pm$ 220 & 4370 $\pm$ 220 & \textbf{3080 $\pm$ 200} & 4310 $\pm$ 220 & 4300 $\pm$ 230 & \textbf{3460 $\pm$ 210} & \textbf{3410 $\pm$ 230} \\ \midrule
     \multirow{2}{*}{Iron} & SR$\uparrow$ & 19.8 $\pm$ 3.7 & 19.4 $\pm$ 3.6 & 54.4 $\pm$ 4.5 & \textbf{74.0 $\pm$ 3.9} & 15.9 $\pm$ 3.2 & 11.6 $\pm$ 2.8 & 56.7 $\pm$ 4.4 & \textbf{66.0 $\pm$ 4.0}\\
     & AS$\downarrow$& 21310 $\pm$ 570 & 21110 $\pm$ 600 & 14860 $\pm$ 800 & \textbf{10800 $\pm$ 770} & 21460 $\pm$ 560 & 22520 $\pm$ 400 & 14100 $\pm$ 790 & \textbf{12200 $\pm$ 740}\\ \midrule
       \multirow{2}{*}{Gold} & SR$\uparrow$ & 7.5 $\pm$ 2.7 & 0.00 & 5.6 $\pm$ 3.5 & \textbf{72.3 $\pm$ 6.4} & 0.00 & 0.00 & 5.3 $\pm$ 3.2 & \textbf{72.2 $\pm$ 6.6}\\
     & AS$\downarrow$& 35100 $\pm$ 400 & $\infty$ & 34400 $\pm$ 1000 & \textbf{15200 $\pm$ 1900} & $\infty$ & $\infty$ & 34500 $\pm$ 900 & \textbf{15300 $\pm$ 1900} \\ \midrule
       \multirow{2}{*}{Diamond} & SR$\uparrow$ & 0.00 & 0.00 & 0.00 & \textbf{66.1 $\pm$ 7.0} & 11.5 $\pm$ 3.7 & 3.3 $\pm$ 1.6 & 0.00 & \textbf{31.7 $\pm$ 6.4} \\
     & AS$\downarrow$& $\infty$ & $\infty$ & $\infty$ & \textbf{19700 $\pm$ 1800} & 34600 $\pm$ 500 & 35940 $\pm$ 40 & $\infty$ & \textbf{27900 $\pm$ 1700} \\ \midrule
       \multirow{2}{*}{Redstone} & SR$\uparrow$ & 0.00 & 0.00 & 11.7 $\pm$ 4.5 & \textbf{49.4  $\pm$ 7.4} & 0.00 & 0.00 & 12.7 $\pm$ 4.5 & \textbf{47.5 $\pm$ 7.0} \\
     & AS$\downarrow$ & $\infty$ & $\infty$ & 32700 $\pm$ 1300 & \textbf{22300 $\pm$ 2100}& $\infty$ &$\infty$ & 32700 $\pm$ 1200 & \textbf{25600 $\pm$ 2000} \\ \midrule
       \multirow{2}{*}{Armor} & SR$\uparrow$ & 23.9 $\pm$ 4.4 & 26.7 $\pm$ 4.4 & 31.8 $\pm$ 4.7 & \textbf{54.4 $\pm$ 5.0} & 23.6 $\pm$ 4.2 & 6.2 $\pm$ 2.5 & 35.4 $\pm$ 4.8 & \textbf{55.6 $\pm$ 4.8} \\
     & AS$\downarrow$ & 29100 $\pm$ 1300 & 28300 $\pm$ 1300 & 26600 $\pm$ 1400 & \textbf{20700 $\pm$ 1400} & 29800 $\pm$ 1200 & 34200 $\pm$ 700 & 25700 $\pm$ 1400 & \textbf{20000 $\pm$ 1400} \\ \midrule
    \multicolumn{10}{l}{- \textit{success rate (SR), average step (AS), $\infty$ (failed after reaching the max steps).} }\\ \bottomrule
    \end{tabular}}
\end{table*}

\subsection{Hyperparameter Analysis of $k$}
Here, we analyze the effect of $k$ when retrieving candidates to match a given query to a goal in our GoG. In test tasks, all of the text instructions of the tasks are provided in a similar structure in the form of ``\textit{$<$verb$>$} \textit{$<$item$>$}'' with a limited number of verbs, such as ``craft a wooden pickaxe''. Therefore, in order to increase the diversity of the task instructions for this analysis, we use GPT-4o-mini to generate 10 rewordings for each of the 66 task instructions used in our experiments. Then, for each of the 660 generated instructions, we retrieve the top-$k$ goals from our GoG, and use an LLM to select the best match from the retrieved goals. In the case of $k=0$, no retrieval is performed and the LLM is directly asked to determine the goal. For $k=1$, the LLM is not required to perform any selection because there is only one option, hence the results for $k=1$ are the same across all LLMs.

The results in Table~\ref{tab:k_exp} show the accuracies of various combinations of $k$ and LLMs in determining the correct goal by checking that the postcondition of the selected goal matches the task. First, we observe that when $k>1$, our goal inference clearly improves the retrieval of matching goals compared to the case of $k=0$, where the LLM must rely solely on its own knowledge to infer the postconditions for the query task. Notably, Gemma 3 exhibited the lowest performance at $k=0$, but its accuracy increased significantly, approaching that of other LLMs, when given access to our knowledge base. Second, we find that increasing $k$ leads to only a marginal improvement in query accuracy, highlighting the robustness of our goal matching strategy, which remains effective even with a small number of retrieved goals.

\begin{table}[!t]
    \centering
    \scalebox{0.85}{
    \begin{tabular}{c | c | c | c }
        \toprule
        $k$  & Gemma 3 & Qwen 2.5 VL & Llama 3.2  \\
        \midrule
         0 & .87 $\pm$ .025 & .93 $\pm$ .020 & .95 $\pm$ .016 \\
         1 & .97 $\pm$ .013 & .97 $\pm$ .013 & .97 $\pm$ .013 \\
         2 & \textbf{.985 $\pm$ .0093} & \textbf{.988 $\pm$ .0084} & \textbf{.988 $\pm$ .0084} \\
         3 & \textbf{.988 $\pm$ .0084} & \textbf{.989 $\pm$ .0078} & \textbf{.987 $\pm$ .0089} \\
         4 & \textbf{.98 $\pm$ .010} & \textbf{.986 $\pm$ .0089} & \textbf{.986 $\pm$ .010} \\
         5 & \textbf{.988 $\pm$ .0084} & \textbf{.995 $\pm$ .0052} & \textbf{.985 $\pm$ .0093} \\
        \bottomrule
    \end{tabular}}
    \caption{Accuracies for different values of $k$ using various LLMs on goal inference for our retrieval method. Confidence intervals are calculated using an alpha of 0.05. Bolded numbers are the significantly best results in each column as determined by pairwise t-tests.}
    \label{tab:k_exp}
\end{table}

When $k=0$, corresponding to the baseline methods that rely solely on internal knowledge to infer the next steps, a significant question arises: why does performance vary considerably in our main experiments between GoG and baselines, even though retrieval performance remains relatively consistent across models? We attribute this performance gap to the recursive retrieval of subgoals in our algorithm, which explicitly constructs a reasoning path, something the baselines lack. However, there is still room for improvement in our approach, which motivates us to further explore how to enable deeper, more structured reasoning over tasks. 

\subsection{Plan Quality Ablation}
\label{sec:plan_ablation}

\begin{table*}[!ht]
    \centering
    \scalebox{0.8}{
    \begin{tabular}{l c c c c c c}
        \toprule
         \multirow{2}{*}{LLM} & Goal & Material & Goal & \multirow{2}{*}{Soundness} & \multirow{2}{*}{Completeness} & \multirow{2}{*}{Efficiency} \\
         & Info & List & Satisfaction & & & \\
        \midrule
        \multirow{4}{6em}{Llama 3.2 Vision} & \xmark & \xmark & .32 $\pm$ .12 & .79  $\pm$ .05 & .82 $\pm$ .045 & .36 $\pm$ .11 \\
        & \cmark & \xmark & .14 $\pm$ .086 & .38 $\pm$ .081 & .33 $\pm$ .089 & .14 $\pm$ .086 \\
        & \xmark & \cmark & \textbf{.95 $\pm$ .052} & \textbf{.99 $\pm$ .022} & \textbf{.99 $\pm$ .016} & \textbf{.95 $\pm$ .052} \\
        & \cmark & \cmark & \textbf{1.0 $\pm$ 0.0} & \textbf{1.0 $\pm$ 0.0} & \textbf{1.0 $\pm$ 0.0} & \textbf{1.0 $\pm$ 0.0} \\
        \midrule
        \multirow{4}{6em}{Qwen 2.5 VL} & \xmark & \xmark & .26 $\pm$ .11 & .77 $\pm$ .043 & .79 $\pm$ .073 & .29 $\pm$ .10 \\
        & \cmark & \xmark & .22 $\pm$ .10 & .76 $\pm$ .055 & .75 $\pm$ .061 & .23 $\pm$ .10 \\
        & \xmark & \cmark & \textbf{.97 $\pm$ .043} & \textbf{.995 $\pm$ .0064} & \textbf{.997 $\pm$ .0043} & \textbf{.953 $\pm$ .052} \\
        & \cmark & \cmark & \textbf{.98 $\pm$ .031} & \textbf{.998 $\pm$ .0034} & \textbf{.998 $\pm$ .0035} & \textbf{.98 $\pm$ .031} \\
        \midrule
        \multirow{4}{6em}{Gemma 3} & \xmark & \xmark & .32 $\pm$ .12 & .82 $\pm$ .038 & .83 $\pm$ .044 & .34 $\pm$ .11 \\
        & \cmark & \xmark & .37 $\pm$ .12 & .78 $\pm$ .060 & .81 $\pm$ .058 & .34 $\pm$ .11 \\
        & \xmark & \cmark & \textbf{.95 $\pm$ .053} & \textbf{.991 $\pm$ .0095} & \textbf{.994 $\pm$ .0072} & \textbf{.88 $\pm$ .082} \\
        & \cmark & \cmark & .83 $\pm$ .094 & .87 $\pm$ .070 & .91 $\pm$ .051 & \textbf{.83 $\pm$ .093} \\
        \bottomrule
    \end{tabular}}
    \caption{Plan quality ablation results, where bolded numbers show the significantly best results in each column as determined by pairwise t-tests for each LLM using an alpha of 0.05.}
    \label{tab:plan_quality_exp}
\end{table*}

Remember that two main components are extracted in our inference stage: the goal tree and the materials and tools list in Section \ref{sec:goal_selection}. In this study, we analyze the effects of the two main components of the prompt for plan generation, and subsequently evaluate the robustness of GoG to imperfect or noisy procedural knowledge by constructing goal graphs from less structured, free-form textual sources. We generate a plan for each of the 66 tasks for each LLM in the ablation study, and alter the information given in the prompt based on the components removed according to the corresponding variant. We experiment with 4 variants: full context information, goal tree information only, material and tool list only, and neither.

\subsubsection{Plan Quality Metrics}

To assess the quality of the generated plans, we use the following metrics, which are based on classical planning literature~\citep{ghallab2004automated}:

\textbf{Goal Satisfaction} assesses overall plan quality by measuring whether a plan, when executed, can possibly achieve the overall task. This means that the plan must include steps to obtain all required tools and materials, and the ordering of the steps should be such that preconditions of each step are not violated. We assign a satisfaction score $g \in \{0,1\}$ to a given plan, with 0 representing a plan that cannot achieve the given task.

\textbf{Soundness} checks that each step of a plan is formulated correctly and is executed with all preconditions being satisfied. For example, ``mine a wooden sword'' would be an invalid plan step. A plan is assigned a soundness score $s \in \{0,1\}$, where 0 indicates that at least one step is invalid. Soundness is an upper bound on goal satisfaction; a plan cannot satisfy a goal if it is not sound. However, a sound plan does not necessarily mean that it will satisfy the goal.

\textbf{Completeness} measures the proportion of materials and tools needed to achieve a task obtained by a plan. More formally, the score is assigned as:
\begin{equation}
c = \min\left(\frac{n_{\text{obtained}}}{n_{\text{needed}}}, 1\right).
\end{equation}

This means that if a plan obtains more materials than necessary, it still obtains a completeness score of 1.

\textbf{Efficiency} determines whether the plan contains more steps than required. A plan may have high goal satisfaction, soundness, and completeness if it obtains many more materials than required, but this would not mean that it is a good plan. Therefore, efficiency checks that a plan consists of only necessary steps, measured as: 
\begin{equation}
    e = \begin{cases}
        \frac{s_{\text{minimal}}}{s_{\text{plan}}} & \text{if} s_{\text{plan}} \geq s_{\text{minimal}}, \\
        0 & \text{otherwise}.
    \end{cases}
\end{equation}
The efficiency of a plan is 0 if the number of steps in the plan $s_{\text{plan}}$ is less than the minimum steps required $s_{\text{minimal}}$ because that would mean that the plan would fail.

To assign scores using these metrics, we use the goal tree retrieved from our knowledge base for a given task. From the goal tree, we calculate the required list of items to complete the task, and the order that the items should be obtained using the preconditions of the goals. We use the goal names in the tree to determine correct wordings, as the goal names are in the form of ``$<$action$>$ $<$item$>$''.

\subsubsection{Plan Quality Ablation Results}

The results are presented in Table~\ref{tab:plan_quality_exp}. The material list refers to the set of materials required to achieve the target goal, and its inclusion significantly improves all four evaluation scores, particularly in terms of using the correct materials, in the correct order, and with greater efficiency across all three LLMs. Overall, providing both goal information and the material list enables the LLM to generate higher-quality plans, as the incorporation of the goal tree, a topological structure that captures goal dependencies, further enhances multi-step reasoning. An exception is observed with Gemma 3, where adding the goal tree slightly reduces performance, likely due to difficulty handling hierarchical input or increased context length from overlapping information with the material list. We leave further investigation of this issue to future work.

\subsubsection{Robustness to Noise}

To assess the robustness of GoG to imperfect or less structured procedural knowledge, we conduct an additional ablation in which the GoG is constructed from noisier inputs. Specifically, instead of directly using structured recipe files, we convert the recipe information into natural language descriptions using an LLM (Qwen2.5 32b)
and merge these descriptions with the corresponding Wiki pages. A new GoG is then constructed from this combined free-form text using the same method as our original GoG. This setting simulates a more realistic scenario in which procedural knowledge is only partially structured, incomplete, or implicitly expressed in text, rather than explicitly encoded as clean recipe graphs.

\begin{table*}[!t]
    \centering
    \scalebox{0.8}{
    \begin{tabular}{l c c c c c}
        \toprule
         \multirow{2}{*}{LLM} & \multirow{2}{*}{Noise} & Goal & \multirow{2}{*}{Soundness} & \multirow{2}{*}{Completeness} & \multirow{2}{*}{Efficiency} \\
         & & Satisfaction & &  & \\
        \midrule
        \multirow{2}{6em}{Llama 3.2 Vision} & \cmark & .97 $\pm$ .043  & .995 $\pm$ .0075  & .996 $\pm$ .0054 & .96 $\pm$ .043 \\
         & \xmark & 1.0 $\pm$ 0.0 & 1.0 $\pm$ 0.0 & 1.0 $\pm$ 0.0 & 1.0 $\pm$ 0.0 \\
        \midrule
        \multirow{2}{6em}{Qwen 2.5 VL} & \cmark & .95 $\pm$ .052 & .993 $\pm$ .0082 & .994 $\pm$ .0063 & .95 $\pm$ .052 \\
         & \xmark & .98 $\pm$ .031 & .998 $\pm$ .0034 & .998 $\pm$ .0035 & .98 $\pm$ .031 \\
        \midrule
        \multirow{2}{6em}{Gemma 3} & \cmark & .66 $\pm$ .12 & .77 $\pm$ .082 & .84 $\pm$ .061 & .66 $\pm$ .12 \\
         & \xmark & .83 $\pm$ .094 & .87 $\pm$ .070 & .91 $\pm$ .051 & .83 $\pm$ .093 \\
        \bottomrule
    \end{tabular}}
    \caption{Robustness to noise. Confidence intervals are calculated using an alpha of 0.05. Noise \cmark{} indicates a GoG constructed from Wiki pages augmented with LLM-generated natural language recipes. Noise \xmark{} corresponds to the original GoG; results from Table~\ref{tab:plan_quality_exp} are reproduced here for direct comparison.}
    \label{tab:noise_robustness_exp}
\end{table*}

Table~\ref{tab:noise_robustness_exp} reports the plan quality metrics under this noisy GoG construction, compared against the original GoG. Across all three LLMs, we observe a consistent degradation in performance when noise is introduced, with the magnitude varying by model. The largest drops occur in efficiency and goal satisfaction, indicating that noise primarily affects the model’s ability to reason about precise preconditions and quantities, rather than completely breaking plan validity. Soundness and completeness remain relatively high, suggesting that the overall goal structure extracted by GoG is largely preserved even under noisy knowledge extraction.

A qualitative inspection of the noisy GoG reveals two dominant sources of error. First, certain goals omit implicit preconditions that are not explicitly stated in free-form text (e.g., missing fuel requirements for smelting), leading to plans that are correct in structure but insufficient in resources. Second, noisy extraction can distort postconditions, such as overestimating the quantity of items produced by a mining action, which in turn affects downstream material calculations. Despite these imperfections, the resulting plans remain largely executable and coherent, highlighting that GoG degrades gracefully rather than catastrophically when structured knowledge is imperfect.

Overall, this experiment demonstrates that while GoG benefits most from clean, semi-structured procedural resources, it is not brittle to moderate noise in knowledge extraction. Instead, its explicit goal dependency structure provides a degree of robustness that allows LLMs to maintain high-quality multi-step reasoning even when procedural information is incomplete or inconsistently specified.

\subsection{Use-case Demonstration}
To illustrate why our method performs better than the baselines, we compare the plans generated by our method and the baselines on the ``craft a diamond axe'' task using Llama 3.2-Vision, as shown in Table~\ref{tab:diamond_axe_plans}. Two types of errors are observed from the plans generated by the baselines. 

\begin{table}[]
    \centering
    \begin{subtable}[T]{.49\linewidth}
        \scalebox{0.48}{
        \begin{tabular}{c|c|c|c}
        \toprule
        Step & Instruction & Target Item & Amount\\
        \hline
        0&chop a tree&logs&4\\
        1&craft planks&planks&12\\
        2&craft stick&stick&8\\
        3&craft crafting table&crafting table&1\\
        4&craft wooden pickaxe&wooden pickaxe&1\\
        5&equip wooden pickaxe&wooden pickaxe&1\\
        6&dig down and break down cobblestone&cobblestone&11\\
        7&craft stone pickaxe&stone pickaxe&1\\
        8&equip stone pickaxe&stone pickaxe&1\\
        9&dig down and break down iron ore&iron ore&3\\
        10&craft a furnace&furnace&1\\
        11&smelt iron ingot&iron ingot&3\\
        12&craft an iron pickaxe&iron pickaxe&1\\
        13&equip iron pickaxe&iron pickaxe&1\\
        14&dig down and break down diamond ore&diamond&3\\
        15&craft diamond axe&diamond axe&1  \\ \bottomrule
        \end{tabular}}
        \subcaption{GoG: Successful plan.}
        \label{tab:diamond_axe_ours_plan}
    \end{subtable}
    \hspace{\fill}
    \begin{subtable}[T]{.49\linewidth}
        \scalebox{0.48}{
        \begin{tabular}{c|c|c|c}
        \toprule
        Step & Instruction & Target Item & Amount\\
        \hline
        0 &chop a tree&logs&9\\
        1 &craft planks&planks&27\\
        2 &craft stick&stick&8\\
        3 &craft crafting table&crafting table&1\\
        4 &craft wooden pickaxe&wooden pickaxe&1\\
        5 &equip wooden pickaxe&wooden pickaxe&1\\
        6 &dig down and break down cobblestone&cobblestone&19\\
        7 &craft stone pickaxe&stone pickaxe&1\\
        8 &equip stone pickaxe&stone pickaxe&1\\
        9 & craft furnace&furnace&1\\
        10 &dig down and break down iron ore&iron ore&3\\
        11 &smelt iron ore&iron ingot&3\\
        12 &craft iron pickaxe&iron pickaxe&1\\
        13 &equip iron pickaxe&iron pickaxe&1\\
        14 &dig down and mine diamond&diamond ore&3\\
        15 &{\color{red}smelt} diamond&diamond&3\\
        16 &craft diamond axe&diamond axe&1\\ \bottomrule
        \end{tabular}}
        \subcaption{HKG: At step 15, the plan tries to ``smelt diamond'', which is unnecessary and invalid.}
        \label{tab:diamond_axe_hkg_plan}
    \end{subtable}
    \begin{subtable}[T]{.49\linewidth}
        \scalebox{0.48}{
        \begin{tabular}{c|c|c|c}
        \toprule
        Step & Instruction & Target Item & Amount\\
        \hline
        0&chop a tree&logs&5\\
        1&craft planks&planks&15\\
        2&craft stick&stick&{\color{red}3}\\
        3&craft crafting table&crafting table&1\\
        4&craft wooden pickaxe&wooden pickaxe&1\\
        5&equip wooden pickaxe&wooden pickaxe&1\\
        6&dig down and break down cobblestone&cobblestone&11\\
        7&craft stone pickaxe&stone pickaxe&1\\
        8&equip stone pickaxe&stone pickaxe&1\\
        9&craft furnace&furnace&1\\
        10&dig down and break down iron ore&iron ore&2\\
        11&smelt iron ore&iron ingot&2\\
        12&chop tree&logs&2\\
        13&craft planks&planks&2\\
        14&craft sticks&stick&2\\
        15&craft wooden pickaxe&wooden pickaxe&1\\
        16&mine iron ore&iron ore&1\\
        17&smelt iron ore&iron ingot&1\\
        18&craft iron pickaxe&iron pickaxe&1\\
        19&equip iron pickaxe&iron pickaxe&1\\
        20&dig down and break down diamond ore&diamond ore&3\\
        21&craft diamond axe&diamond axe&1  \\ \bottomrule
        \end{tabular}}
        \subcaption{GraphRAG: Contains replanning steps because the plan failed to obtain enough sticks in step 3.}
        \label{tab:diamond_axe_graphrag_plan}
    \end{subtable}
    \hspace{\fill}
    \begin{subtable}[T]{.49\linewidth}
        \scalebox{0.48}{
        \begin{tabular}{c|c|c|c}
        \toprule
        Step & Instruction & Target Item & Amount\\
        \hline
        0&chop a tree&logs&5\\
        1&craft planks&planks&16\\
        2&craft stick&stick&{\color{red}4}\\
        3&craft crafting table&crafting table&1\\
        4&craft wooden pickaxe&wooden pickaxe&1\\
        5&equip wooden pickaxe&wooden pickaxe&1\\
        6&dig down and break down cobblestone&cobblestone&11\\
        7&craft stone pickaxe&stone pickaxe&1\\
        8&equip stone pickaxe&stone pickaxe&1\\
        9&craft furnace&furnace&1\\
        10&dig down and break down iron ore&iron ore&2\\
        11&smelt iron ore&iron ingot&2\\
        12&chop tree&logs&2\\
        13&craft planks&planks&2\\
        14&craft sticks&sticks&2\\
        15&craft stone pickaxe&stone pickaxe&1\\
        16&mine iron ore&iron ore&1\\
        17&smelt iron ore&iron ingot&1\\
        18&craft iron pickaxe&iron pickaxe&1\\
        19&equip iron pickaxe&iron pickaxe&1\\
        20&dig down and break down diamond ore&diamond ore&3\\
        21&craft diamond axe&diamond axe&1\\  \bottomrule
        \end{tabular}}
        \subcaption{Vanilla: Contains replanning steps because the plan failed to obtain enough sticks in step 3. }
        \label{tab:diamond_axe_vanilla_plan}
    \end{subtable}
    \caption{Plans generated using various methods for the ``craft a diamond axe'' task. Text in red highlights the key errors in the generated plans.}
    \label{tab:diamond_axe_plans}
\end{table}

First, we observe hallucinations in intermediate planning steps. For example, the plan generated by HKG includes the step ``smelt diamond,'' in Table \ref{tab:diamond_axe_hkg_plan}, instead of the correct step, ``smelt diamond \textbf{ore}''. This semantic distinction, although seemingly minor, is important because the ``smelt'' action should be applied to the object ``diamond ore'', not the final object ``diamond''. Additionally, in this scenario, mining diamond with an iron pickaxe would directly give the agent diamond and not diamond ore, and thus ``smelt diamond ore'' would not be an applicable step. This is unlike many other ores in the game, such as iron and gold, which do need to be smelted after being mined. We hypothesize that this error stems from the LLM’s lack of understanding of game mechanics. Although HKG supplies a list of materials and tools, this information alone is insufficient for constructing a valid plan. In contrast, our top-down approach provides structured goal decomposition, which better aligns with the procedural nature of in-game tasks. 

Second, both GraphRAG and the Vanilla baseline exhibit similar shortcomings in long-horizon planning. In particular, both generate plans that produce too few sticks, forcing the agent to replan and ultimately fail to complete the task, as shown in Tables \ref{tab:diamond_axe_graphrag_plan} and \ref{tab:diamond_axe_vanilla_plan}. This limitation arises from the LLM’s inability to reason about quantities, as successfully completing the task requires both knowledge of crafting recipes and the capacity to compute the required number of items based on those recipes.

Our GoG is constructed from practical Minecraft resources, such as the Minecraft Wiki and in-game crafting recipes, which provide rich information about both crafting procedures and the quantities of required materials. Unlike the baselines, GoG not only models goal-to-goal dependencies, but also explicitly encodes preconditions and postconditions for each goal. For example, ``1 crafting table, 2 sticks, and 3 planks'' for crafting a wooden pickaxe. This structured representation significantly mitigates the two types of errors identified in our analysis, as shown in Table~\ref{tab:diamond_axe_ours_plan}.

\section{Conclusion}
In this work, we introduced Goal-Oriented Graphs (GoGs), a novel framework for structuring and retrieving procedural knowledge to enhance long-horizon reasoning in LLM-based agents. Unlike entity-centric approaches such as GraphRAG, which often fragment information into isolated triples, GoGs represent goals and their dependencies in a hierarchical structure that enables coherent retrieval of reasoning chains. Experiments in Minecraft, a rich and open-ended environment requiring hierarchical planning and supported by semi-structured procedural knowledge, demonstrated that GoG substantially improves reasoning performance and task success over existing retrieval-augmented baselines.

Our focus on Minecraft provided a complex yet controlled testbed to evaluate multi-step reasoning and goal decomposition. Pretrained LLMs may already possess partial knowledge of Minecraft, which could influence extraction and planning performance. However, the consistent performance gap between GoG and vanilla LLM baselines, which both rely on the same pretrained model, suggests that the improvement stems from structured goal reasoning rather than memorized domain knowledge. Nonetheless, we recognize that assessing GoG in domains unfamiliar to pretrained LLMs remains an important direction for future work. To this end, we plan to evaluate GoG in synthetic or obfuscated environments (e.g., randomized item names) to examine robustness against prior-knowledge leakage and test its adaptability to entirely new contexts.

Another key direction for future research is extending GoG beyond structured procedural corpora. While Minecraft’s recipe and Wiki data offer semi-structured goal hierarchies, this property is not unique to games; many real-world domains (e.g., cooking, manufacturing, troubleshooting, medical workflows) contain similarly structured instructional or procedural texts. GoG provides a principled way to formalize and exploit these common goal-oriented data sources. At the same time, our framework can extract latent hierarchies from unstructured text, although, as shown in our noise ablation, performance depends strongly on textual clarity and the extent to which procedural relations are implicitly recoverable.

A key limitation of GoG is its reliance on external procedural knowledge that can be reasonably extracted into goal–precondition–effect structures. In domains where such knowledge is absent, highly implicit, or fundamentally non-symbolic, the benefits of GoG may be reduced. Our noise ablation suggests that GoG degrades gracefully under moderate imperfections, but fully unstructured environments remain an open challenge.

However, GoG provides a foundation for handling creative or abstract goals that lack explicit recipes. While such tasks fall outside the current evaluation scope, their underlying reasoning process, involving breaking down a high-level creative objective into actionable subgoals, aligns naturally with the GoG paradigm. Future extensions could leverage the LLM’s generative capabilities to dynamically construct and refine GoGs for novel tasks on the fly, blending retrieved procedural knowledge with generative reasoning to support open-ended creativity.

We also plan to expand GoG in several technical directions.
First, while the current implementation retrieves a single, coherent reasoning chain, many tasks admit multiple valid subgoal decompositions (e.g., crafting items through different resource routes). In this work, we focus on evaluating plan feasibility, correctness, and material reasoning rather than global optimality with respect to cost or step count. Accordingly, GoG retrieves a single coherent reasoning chain, and our evaluation metrics assess whether this chain is executable and sufficient to achieve the goal. Quantitative analysis of path diversity, alternative goal chains, and optimality-aware retrieval (e.g., cost- or success-weighted edges) remains an important direction for future work.
Second, although the present study emphasizes offline graph construction, GoG naturally supports incremental online replanning. As an agent executes a plan and observes environment feedback, achieved or infeasible goals can be dynamically updated or replaced within the graph, allowing continuous refinement of reasoning chains in real time.
Finally, we will investigate scalability and error mitigation strategies, including goal clustering, semantic de-duplication, and graph sparsification, to ensure efficient retrieval and minimize noise in large-scale or cross-domain settings.

Overall, GoG offers an interpretable and extendable framework for goal-centric reasoning in domains where procedural knowledge can be expressed or recovered as goal dependencies, and complements existing retrieval-augmented and planning-based approaches. While validated here in Minecraft, the principles underlying GoG, namely the explicit goal dependency modelling and recursive goal retrieval, are broadly applicable across domains where procedural reasoning is key and where external procedural knowledge can be reasonably extracted. We hope this work contributes a meaningful step toward bridging structured planning and LLM-based reasoning, and lays the foundation for future agents capable of dynamically learning, adapting, and reasoning about goals in complex, open-ended environments.

\section*{Acknowledgements}
This research is supported by the RIE2025 Industry Alignment Fund – Industry Collaboration Projects (IAF-ICP) (Award I2301E0026), administered by A*STAR, as well as supported by Alibaba Group and NTU Singapore through Alibaba-NTU Global e-Sustainability CorpLab (ANGEL).




\bibliographystyle{elsarticle-num-names}
\bibliography{ref}

\clearpage
\appendix

\section{Prompts}
\label{appendix:prompts}

In this section, we provide various prompts used in our experiments. Figure~\ref{fig:goal_extraction_prompt} contains the prompt for goal extraction used for the construction of GoG. Figure~\ref{fig:goal_inference_prompt} contains the prompt used to determine which goal the agent should pursue. Figure~\ref{fig:planning_prompt} contains the prompt used to generate the plan. Figure~\ref{fig:graphrag_prompt} contains the prompt used to extract entities and relationships from the source text (Minecraft Wiki pages and recipe files), which is used to build the knowledge graph used by GraphRAG. Text in curly braces represent placeholders that should be replaced by contextual information at inference time.

\begin{figure}
    \centering
    \begin{subfigure}[T]{0.49\textwidth}
        \centering
        \scalebox{0.5}{
        \begin{tcolorbox}[
          colback=white, colframe=black, arc=3mm, width=2\linewidth,
          title=\textbf{Prompt for Extracting Goals and Subgoals, Part 1}, 
          coltitle=white, colbacktitle=gray, fonttitle=\bfseries
        ]
-Goal-
    
Given a portion of a document and relevant in-game recipes about the game Minecraft, extract actionable in-game goals that a player can achieve. Use only the content from the given document and in-game recipe JSONs to construct goals and subgoals. Do not infer or add goals beyond what is explicitly described. Focus solely on the core Minecraft experience. Exclude any content related to Minecraft spinoff games (e.g., Minecraft Dungeons, Minecraft Legends).

-Steps-

1. Identify relevant goals that a player can achieve in the game. For each goal, extract the following attributes:

- name: Name of the goal. Use short, specific names in the form of ``$<$action$>$ $<$minecraft\_item$>$'', such as ``craft planks'', ``mine cobblestone'', or ``smelt charcoal''. For tools with different grades such as ``wooden'' or ``stone'', use ``$<$action$>$ $<$grade$>$ $<$minecraft\_tool$>$'', such as ``craft a wooden pickaxe'' or ``craft a stone sword''.

- description: A concise explanation of what the goal entails.

- req\_tools: Needed tools to complete the goal, as a JSON object where keys are Minecraft tools and values are 1. For tools with multiple grades (e.g. wooden or stone), specify the tool grade and only include the lowest grade needed. Crafting tables and furnaces are considered as tools, and their usage can be determined by document text, recipes, and summaries. Smelting using a furnace always requires ``fuel'' as a tool. Use ``None'' (just as a standalone string, not as a JSON object or set or list) if no tools are needed.

- req\_materials: Needed materials to complete the goal, as a JSON object where keys are Minecraft items and values are needed quantities of that item. If no materials are needed, set this to ``None'' (just as a standalone string, not as a JSON object or set or list).

- postconditions: The resulting state or item after completing the goal, as a JSON object where the keys are Minecraft items and values are the quantity. If there are no post-conditions, set this to ``None'' (just as a standalone string, not as a JSON object or set or list).

Before writing each goal, generate reasoning as to where the information about the goal comes from. If it comes from a shaped crafting recipe, you must use the format as described above, otherwise write a brief sentence.

Format each goal as a tuple: 

(``goal''\{tuple\_delimiter\}``$<$name$>$''\{tuple\_delimiter\}

``$<$description$>$''\{tuple\_delimiter\}``$<$req\_tools$>$''\{tuple\_delimiter\}

``$<$req\_materials$>$''\{tuple\_delimiter\}``$<$postconditions$>$'')

... (continued in Figure~\ref{fig:goal_extraction_prompt_p2}) ...
        \end{tcolorbox}}
        \caption{}
        \label{fig:goal_extraction_prompt_p1}
    \end{subfigure}%
    \hspace{\fill}%
    \begin{subfigure}[T]{0.49\textwidth}
        \centering
        \scalebox{0.5}{
        \begin{tcolorbox}[
          colback=white, colframe=black, arc=3mm, width=2\linewidth,
          title=\textbf{Prompt for Extracting Goals and Subgoals, Part 2}, 
          coltitle=white, colbacktitle=gray, fonttitle=\bfseries
        ]
2. From the goals identified in step 1, identify subgoals that are needed for the achievement of the goal. For every goal, establish subgoal relationships between the goal and associated subgoals for each required material and tool that must be obtained or crafted, as identified by $<$req\_tools$>$ or $<$req\_materials$>$.
For each goal-subgoal relationship, extract the following information:

- goal\_name: Name of the higher-level goal, which must exist in the goals identified in step 1.

- subgoal\_name: Name of the subgoal that is used by the goal.

- relationship\_description: Explanation as to how and why the higher-level goal and the subgoal are related to each other.

Format each relationship as a tuple:

(``subgoal''\{tuple\_delimiter\}``$<$goal\_name$>$''\{tuple\_delimiter\}

``$<$subgoal\_name$>$''\{tuple\_delimiter\}``$<$relationship\_description$>$'')

3. Return a single list of tuples of all goals and subgoals as extracted from steps 1 and 2. Use **\{record\_delimiter\}** as the list delimiter. If either tools or materials are ambiguous or missing, omit the goal. Do not repeat the same goal in the list.

4. When finished, output \{completion\_delimiter\}.

Only output the list as instructed without any explanation, summary, or other text. If there is no relevant information in the document, just output \{completion\_delimiter\}.

Here are some examples:

\{examples\}

\#\#\#\#\#\#\#\#\#\#\#\#\#\#\#\#\#\#\#\#\#\#

-Real Data-

\#\#\#\#\#\#\#\#\#\#\#\#\#\#\#\#\#\#\#\#\#\#

Document Text:

\{input\_text\}

--- End of Document ---

Goals and Subgoals:
        \end{tcolorbox}}
        \caption{}
        \label{fig:goal_extraction_prompt_p2}
    \end{subfigure}
    \caption{The prompt used to extract goals and subgoals from source texts to build our knowledge base.}
    \label{fig:goal_extraction_prompt}
\end{figure}

\begin{figure*}
    \centering
     \begin{tcolorbox}[
      colback=white, colframe=black, arc=3mm, width=1\linewidth,
      title=\textbf{Prompt for Goal Inference}, 
      coltitle=white, colbacktitle=gray, fonttitle=\bfseries
    ]
You are a MineCraft game expert and you can guide agents to complete complex tasks. For a given game screen, task, and context information, you need to complete ``goal inference'' and ``visual inference''.

The context information is a set of possible goals to choose from for ``goal inference''.

``goal inference'': According to the task, you need to select the goal from given options that best matches the given query.

``visual inference'': According to the game screen, you need to infer the following aspects: health bar, food bar, hotbar, environment.

\{Examples\}

Here is a game screen and task, you MUST respond in JSON format as shown in the example outputs WITHOUT further explanation, introduction, or extra text. Complete ``goal inference'' by setting it to the value of the ``name'' of the option that best matches the given task as shown in the example. Other fields should be completed based on the given game screen.

$<$task$>$: \{task\}

$<$context$>$:

\{context\}

Output:
    \end{tcolorbox}
    \caption{The prompt used for goal inference for our proposed method. ``Context'' consists of the top-$k$ retrieved goals from the knowledge base.}
    \label{fig:goal_inference_prompt}
\end{figure*}

\begin{figure*}
    \centering
    \scalebox{0.65}{
     \begin{tcolorbox}[
      colback=white, colframe=black, arc=3mm, width=1\linewidth,
      title=\textbf{Prompt for Plan Generation}, 
      coltitle=white, colbacktitle=gray, fonttitle=\bfseries
    ]
You are a MineCraft game expert and you can guide agents to complete complex tasks. For a given overall goal, game screen, hierarchy of goals, and list of needed materials, construct a ordered plan that completes the given task. The goal hierarchy is structured as a JSON object whose keys are names of goals, and values are information about the goal and its and subgoals. You will be given a list of tools and materials and amounts needed for you to obtain and craft to complete the overall goal. Based on the information from the goal hierarchy and the list of tools and materials, create a plan in JSON format as shown in the following example:

\#\#\#\#\#\#\#\#\#\#\#\#\#\#\#\#\#\#\#\#\#\#

-Example-

\#\#\#\#\#\#\#\#\#\#\#\#\#\#\#\#\#\#\#\#\#\#

\{example\}

\#\#\#\#\#\#\#\#\#\#\#\#\#\#\#\#\#\#\#\#\#\#

\#\#\#\#\#\#\#\#\#\#\#\#\#\#\#\#\#\#\#\#\#\#

-Real Task-

\#\#\#\#\#\#\#\#\#\#\#\#\#\#\#\#\#\#\#\#\#\#

$<$goal$>$

\{goal\}

$<$visual info$>$

\{visual\_info\}

$<$goal hierarchy$>$

\{goal\_hierarchy\}

$<$materials and tools$>$

\{materials\_and\_tools\}

$<$planning$>$

Complete $<$planning$>$ for the given overall $<$goal$>$ with valid JSON as instructed and in the format shown in the example. Use the information in the goal hierarchy, game screen, and list of tools and materials and their amounts to generate ``task'' and ``goal'' in each step of the plan. Use the same wording styles and patterns for the ``task'' in each step as shown in the example plan. Only output the plan as a valid JSON object with no additional text, introduction, or explanation. Do not use Markdown.
    \end{tcolorbox}}
    \caption{The prompt used for planning for our proposed method.}
    \label{fig:planning_prompt}
\end{figure*}

\begin{figure*}
    \centering
    \scalebox{0.65}{
     \begin{tcolorbox}[
      colback=white, colframe=black, arc=3mm, width=1\linewidth,
      title=\textbf{Prompt for GraphRAG Entity Extraction}, 
      coltitle=white, colbacktitle=gray, fonttitle=\bfseries
    ]
-Goal-
Given a text document about the game ``Minecraft'' and a list of entity types, identify all game-related entities of those types from the text and all relationships among the identified entities. The document comes from the Minecraft Wiki, and may contain headers, tables, recipes, and text in other formats. Focus on entities related to items, tools, and crafting. You can ignore entities related to game patches and versions, entities outside of the Minecraft game, and abstract entities that are unrelated to gameplay.

-Steps-

1. Identify all game-related entities. For each identified entity, extract the following information:

- entity\_name: Name of the entity, capitalized

- entity\_type: One of the following types: [\{entity\_types\}]

- entity\_description: Comprehensive description of the entity's attributes and activities

Format each entity as 

(``entity''\{tuple\_delimiter\}$<$entity\_name$>$\{tuple\_delimiter\}

$<$entity\_type$>$\{tuple\_delimiter\}$<$entity\_description$>$)

2. From the entities identified in step 1, identify all pairs of (source\_entity, target\_entity) that are *clearly related* to each other.

For each pair of related entities, extract the following information:

- source\_entity: name of the source entity, as identified in step 1

- target\_entity: name of the target entity, as identified in step 1

- relationship\_description: explanation as to why you think the source entity and the target entity are related to each other

- relationship\_strength: a numeric score indicating strength of the relationship between the source entity and target entity

 Format each relationship as
 
 (``relationship''\{tuple\_delimiter\}$<$source\_entity$>$\{tuple\_delimiter\}
 
 $<$target\_entity$>$\{tuple\_delimiter\}$<$relationship\_description\>
 
 \{tuple\_delimiter\}$<$relationship\_strength$>$)

3. Return output in English as a single list of all the entities and relationships identified in steps 1 and 2. Use **\{record\_delimiter\}** as the list delimiter.

4. When finished, output \{completion\_delimiter\}

\#\#\#\#\#\#\#\#\#\#\#\#\#\#\#\#\#\#\#\#\#\#

-Examples-

\#\#\#\#\#\#\#\#\#\#\#\#\#\#\#\#\#\#\#\#\#\#

\{examples\}

\#\#\#\#\#\#\#\#\#\#\#\#\#\#\#\#\#\#\#\#\#\#

-Real Data-

\#\#\#\#\#\#\#\#\#\#\#\#\#\#\#\#\#\#\#\#\#\#

Entity\_types: \{entity\_types\}

Text: \{input\_text\}

\#\#\#\#\#\#\#\#\#\#\#\#\#\#\#\#\#\#\#\#\#\#

Output:
    \end{tcolorbox}}
    \caption{The prompt used for extracting entities and relationships for the construction of the knowledge graph used by GraphRAG. The entity types given to the LLM are: ``item'', ``block'', ``equipment'', ``location'', ``event'', ``npc''.}
    \label{fig:graphrag_prompt}
\end{figure*}

\section{Experimental Tasks}
\label{appendix:tasks}

Table~\ref{tab:exp_tasks} presents the list of tasks for each task group, along with the corresponding maximum number of in-game steps allowed. In total, there are 66 tasks, categorized into seven groups: wood, stone, iron, gold, diamond, redstone, and armor. As expected, more complex tasks require a greater number of steps to complete.

\begin{table*}[]
    \centering
    \scalebox{0.7}{
    \begin{tabular}{c|c|p{11cm}|c}
        \toprule
        Task Group & \#Tasks & Task description & Max Steps \\
        \hline
        Wood & 10 & craft a wooden shovel, craft a wooden pickaxe, craft a wooden axe, craft a wooden hoe, craft a stick, craft a crafting table, craft a wooden sword, craft a chest, craft a bowl, craft a ladder & 2400 \\ \midrule
        Stone & 9 & craft a stone shovel, craft a stone pickaxe, craft a stone axe, craft a stone hoe, smelt a charcoal, craft a smoker, craft a stone sword, craft a furnace, craft a torch & 6000 \\ \midrule
        Iron & 16 & craft a iron shovel, craft a iron pickaxe, craft a iron axe, craft a iron hoe, craft a bucket, craft a hopper, craft a rail, craft a iron sword, craft a shears, craft a smithing table, craft a tripwire hook, craft a chain, craft an iron bars, craft an iron nugget, craft a blast furnace, craft a stonecutter & 24000 \\ \midrule
        Gold & 6 & craft a golden shovel, craft a golden pickaxe, craft a golden axe, craft a golden hoe, craft a golden sword, smelt and craft a gold ingot& 36000 \\ \midrule
        Diamond & 6 & craft a diamond shovel, craft a diamond pickaxe, craft a diamond axe, craft a diamond hoe, craft a diamond sword, craft a jukebox & 36000 \\ \midrule
        Redstone & 6 & craft a piston, craft a redstone torch, craft an activator rail, craft a compass, craft a dropper,
        craft a note block & 36000 \\ \midrule 
        Armor & 13 & craft shield, craft iron chestplate, craft iron boots, craft iron leggings, craft iron helmet, craft diamond helmet, craft diamond chestplate, craft diamond leggings, craft diamond boots,  craft golden helmet, craft golden leggings, craft golden boots, craft golden chestplate & 36000 \\ \bottomrule
    \end{tabular}}
    \caption{Tasks used in our main experiments.}
    \label{tab:exp_tasks}
\end{table*}

\section{GoG Construction}
\label{appendix:construction}

In this section, we provide cost estimations for the construction and inference phases of our proposed GoG framework in Tables~\ref{tab:construction_cost} and \ref{tab:inference_cost}, respectively.

\begin{table}[]
    \centering
    \begin{tabular}{p{0.23\textwidth}|p{0.7\textwidth}}
        \hline
        Step & LLM/Embedding Cost \\
        \hline
        Chunking & N/A \\
        Extracting Goals and Subgoals & One LLM call for each chunk (costing about US\$2-3 in total for 2.7M text source using GPT-4o-mini). \\
        Goal Merge and Subgoal Derivation & Cosine similarity between the embeddings of each goal or sub-goal property (low cost). \\
        \hline
    \end{tabular}
    \caption{GoG construction costs (incurred once).}
    \label{tab:construction_cost}
\end{table}

\begin{table}[]
    \centering
    \begin{tabular}{p{0.2\textwidth}|p{0.7\textwidth}}
        \hline
        Step & LLM/Embedding Cost \\
        \hline
        Query & Embed the task query. \\
        Retrieval & Cosine similarity between the embedded query and each goal in the graph. \\
        Goal Inference & One LLM call at the start of the task, and one LLM call for replanning if the agent does not make progress after $x$ steps. \\
        Planning & One LLM call at the start of the task, and one LLM call for replanning if the agent does not make progress after $x$ steps. \\
        \hline
    \end{tabular}
    \caption{Online inference costs (incurred per task).}
    \label{tab:inference_cost}
\end{table}

\end{document}